# Fine-Tuning Vision-Language Model for Automated Engineering Drawing Information Extraction

Muhammad Tayyab Khan[1, 3], Lequn Chen[2], Ye Han Ng[2], Wenhe Feng[1], Nicholas Yew Jin Tan[1], Seung Ki Moon[3]

[1] Singapore Institute of Manufacturing Technology (SIMTech), Agency for Science, Technology and Research (A*STAR), 5 CleanTech Loop, #01-01 CleanTech Two Block B, Singapore 636732, Republic of Singapore

[2] Advanced Remanufacturing and Technology Centre (ARTC), Agency for Science, Technology and Research (A*STAR), 3 CleanTech Loop, #01-01 CleanTech Two, Singapore 637143, Republic of Singapore

[3] School of Mechanical and Aerospace Engineering, Nanyang Technological University, 639798, Singapore

**Abstract**

Geometric Dimensioning and Tolerancing (GD&T) plays a critical role in manufacturing by defining acceptable variations in part features to ensure component quality and functionality. However, extracting GD&T information from 2D engineering drawings is a time-consuming and labor-intensive task, often relying on manual efforts or semi-automated tools. To address these challenges, this study proposes an automated and computationally efficient GD&T extraction method by fine-tuning *Florence-2*, an open-source vision-language model (VLM). The model is trained on a dataset of 400 drawings with ground truth annotations provided by domain experts. For comparison, two state-of-the-art closed-source VLMs, *GPT-4o* and *Claude-3.5-Sonnet*, are evaluated on the same dataset. All models are assessed using precision, recall, F1-score, and hallucination metrics. Due to the computational cost and impracticality of fine-tuning large closed-source VLMs for domain-specific tasks, GPT-4o and Claude-3.5-Sonnet are evaluated in a zero-shot setting. In contrast, Florence-2, a smaller model with 0.23 billion parameters, is optimized through full-parameter fine-tuning across three distinct experiments, each utilizing datasets augmented to different levels. The results show that Florence-2 achieves a 29.95% increase in precision, a 37.75% increase in recall, a 52.40% improvement in F1-score, and a 43.15% reduction in hallucination rate compared to the best-performing closed-source model. These findings highlight the effectiveness of fine-tuning smaller, open-source VLMs like Florence-2, offering a practical and efficient solution for automated GD&T extraction to support downstream manufacturing tasks.

## 1 Introduction

Geometric Dimensioning and Tolerancing (GD&T) is essential in manufacturing for specifying allowable variations in part features, ensuring that components meet precise functional and quality standards [1]. GD&T uses standardized symbols and rules to clearly communicate geometric requirements on engineering drawings, with Feature Control Frames (FCFs) specifying tolerances, geometric characteristics, and relevant datums. Accurate interpretation of GD&T symbols is crucial for tasks such as inspection, quality control, and assembly, as even minor misinterpretations can lead to production defects and non-conforming parts [2].

Traditionally, extracting GD&T information from 2D engineering drawings has been a manual process, often involving techniques like ballooning, where engineers manually mark and label features [3]. While tools such as Mitutoyo's *MeasurLink* [4] assist in these processes, these manual methods remain slow, error-prone, and difficult to scale in high-volume production environments. Additionally, variability in interpretation and data entry errors can result in costly rework and production delays [5].

Recent advancements in machine learning methods, such as object detection algorithms (YOLO [6]) and Optical Character Recognition (OCR) technologies (Tesseract [7]), have been applied to automate GD&T extraction, reducing manual effort. However, these approaches still face limitations when dealing with complex GD&T features like composite tolerances, modifiers, non-standard text orientations, and requirements for extensive, accurately labeled datasets [5]. Such limitations constrain their use in advanced manufacturing applications, where high precision is critical.

In response to these challenges, vision-language models (VLMs) [8–10] offer an integrated approach by simultaneously processing visual and textual information, thereby enhancing the recognition and interpretation of GD&T symbols and annotations. VLMs can handle complex layouts and symbols that traditional methods struggle with, leveraging efficient querying techniques to deliver reliable results even with smaller labeled datasets [11–13]. Additionally, fine-tuning [14–16], even with limited labeled data, can significantly enhance accuracy, making VLMs well-suited for industrial applications that demand high precision in GD&T extraction.

This study focuses on applying full-parameter fine-tuning technique to Florence-2 [17], an open-source VLM, to enhance its performance for automated GD&T extraction from 2D engineering drawings. To benchmark its performance, two leading closed-source VLMs, GPT-4o [18] and Claude-3.5-Sonnet [19], are evaluated in a zero-shot setting [20–22]. The evaluation uses key metrics such as precision, recall, F1-score, and hallucination, leveraging a dataset of 400 drawings with ground truth annotations. Due to the high computational cost and infeasibility of fine-tuning large closed-source VLMs for domain-specific tasks, this study underscores the effectiveness of fine-tuning smaller, open-source models like Florence-2 as a practical and cost-effective solution for automating GD&T extraction to support downstream manufacturing tasks.

## 2 Methodology Overview

The proposed framework automates GD&T information extraction from 2D engineering drawings by fine-tuning Florence-2, as illustrated in Figure 1. The model is trained on a dataset of 400 annotated drawings, with domain experts providing ground truth annotations for accurate performance evaluation. Text queries are generated for each drawing, and the data is organized into a CSV file linking images with relevant questions, ensuring consistency in VLM input.

For comparison, GPT-4o and Claude-3.5-Sonnet are evaluated in a zero-shot configuration due to the high computational cost and impracticality of fine-tuning large models for domain-specific tasks. In contrast, Florence-2 undergoes full-parameter fine-tuning [14,23,24], benefiting from its smaller size, which allows for more efficient optimization and makes it feasible for domain-specific tasks like GD&T extraction.

The fine-tuning process is conducted using three distinct datasets generated through data augmentation, with each dataset containing one, two, and four queries per image to capture varying levels of feature complexity. This augmentation broadens the model's ability to extract GD&T features across multiple scenarios, enhancing its robustness. After fine-tuning the model on these three datasets, the outputs of the fine-tuned models, along with the outputs from the two closed-source models (GPT-4o and Claude-3.5-Sonnet), are compared against the ground truth annotations. The performance of all models is evaluated using four key metrics: precision, recall, F1-score, and hallucination rate.

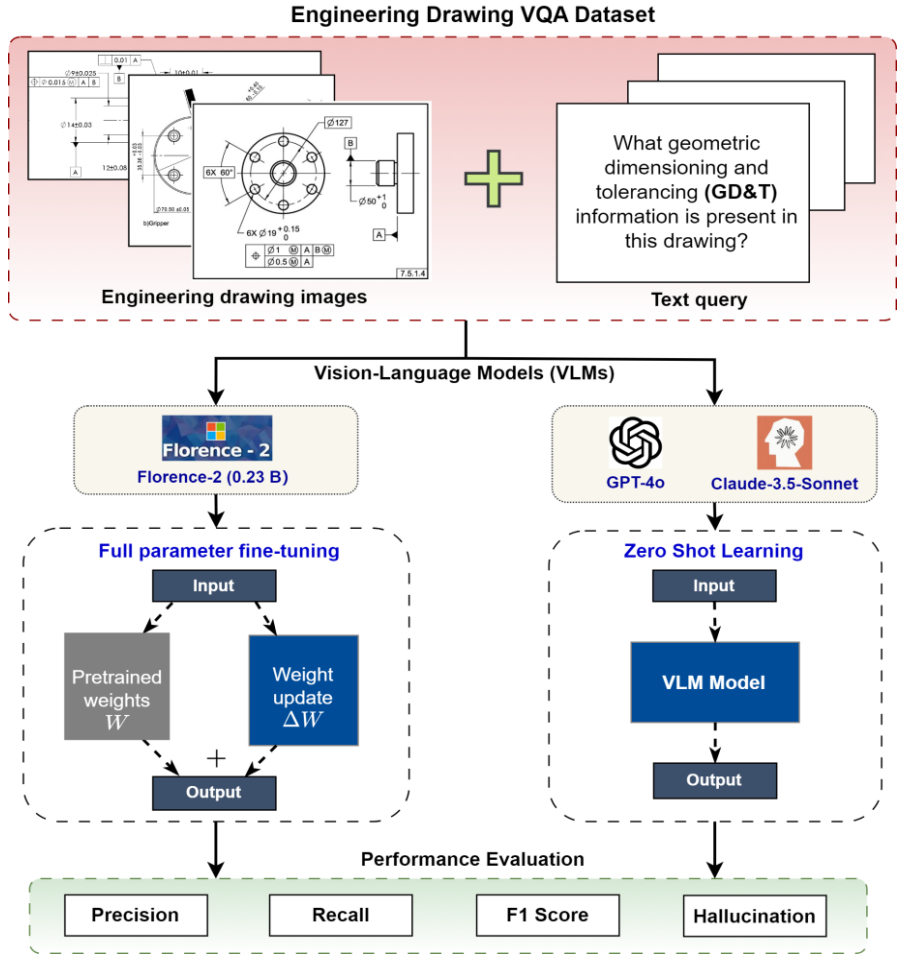

**Fig. 1.** Overview of the proposed framework for GD&T information extraction from 2D engineering drawings.

## 3 Dataset Development

The dataset development process establishes a robust foundation for evaluating VLMs in extracting GD&T information. The dataset consists of 400 annotated 2D engineering drawings, all uniformly converted to PNG format to ensure compatibility across VLM input requirements and improve processing consistency. Drawings originally in other formats, such as PDF or JPEG, are preprocessed to maintain a standardized PNG format.

Each drawing is annotated by domain experts, focusing on three essential GD&T components: geometric characteristics, tolerances, and datum references. These annotations are stored in JSON format, capturing the necessary GD&T details. To improve cross-model compatibility and ease of interpretation, 14 commonly used GD&T symbols are represented using Unicode characters, which resolves challenges in processing complex or uncommon GD&T symbols that may be difficult for language models.

The dataset is structured with both image and text components for compatibility with VLMs. A CSV file links each drawing to a unique index, query questions, and ground truth annotations, facilitating seamless integration with the models' input requirements and allowing precise comparisons with model-generated predictions. Data augmentation techniques are applied to create three distinct datasets:

- *Dataset 1:* 400 base images, each paired with one query question.
- *Dataset 2:* 400 base images, each paired with two different query questions.
- *Dataset 3:* 400 base images, each paired with four different query questions.

These datasets are used to fine-tune Florence-2 and evaluate its performance against the ground truth dataset. Figure 2 illustrates an example of a 2D engineering drawing alongside its corresponding expert-labeled ground truth in JSON format.

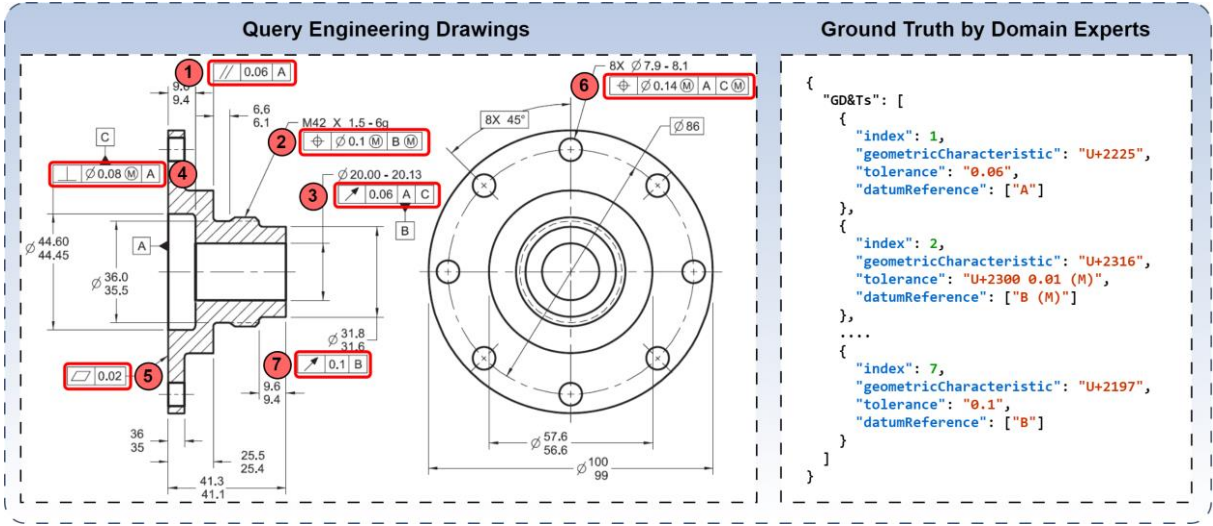

**Fig. 2.** Example of a 2D engineering drawing (*left*) and its corresponding ground truth annotations in JSON format (*right*), illustrating the structure and labeling process.

## 4 Experiments

This section outlines the experimental setup and evaluation process for three selected VLMs (GPT-4o, Claude-3.5-Sonnet, and Florence-2) in extracting GD&T information from 2D engineering drawings.

### 4.1 Model Selection and Data Preparation

The experimental design includes selecting the models and preparing the data for GD&T extraction:

- *GPT-4o and Claude-3.5-Sonnet*: These state-of-the-art closed-source models are evaluated in a zero-shot configuration, meaning they rely solely on their pre-trained knowledge to process the base dataset of 400 images. No additional fine-tuning is performed due to the computational cost and impracticality of adapting these large models to domain-specific tasks.
- *Florence-2-base*: Florence-2, an open-source model with 0.23 billion parameters, is fine-tuned using the three datasets generated through data augmentation (one, two, and four queries per image). Each dataset is split in an 8:2 ratio for training and validation to ensure that the model effectively learns to extract GD&T features across a range of complexities and is validated on unseen data.

The 400 engineering drawings exhibit substantial variability in the number of GD&T entries, ranging from none in some drawings to as many as 14 in others, as depicted in Figure 3. This variability highlights the importance of robust model training to effectively manage diverse GD&T scenarios, enabling the model to generalize well across simple and complex drawings.

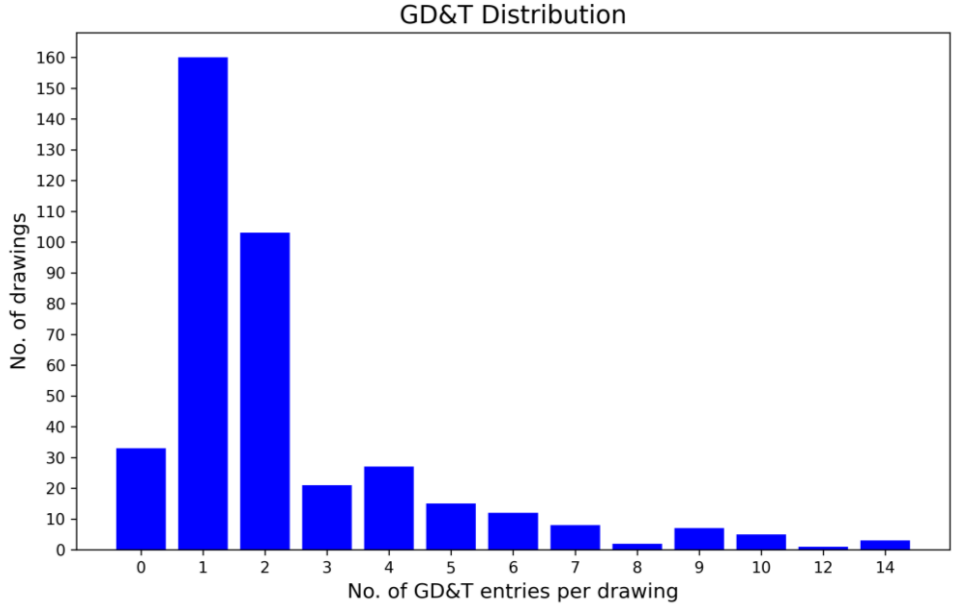

**Fig. 3.** Distribution of GD&T entries across the 400 engineering drawings, illustrating the variability in the GD&T components within the dataset.

### 4.2 Florence-2 Fine-tuning

Florence-2 (with 230 million parameters) is fine-tuned using an NVIDIA GeForce RTX 4090 GPU in a full-parameter tuning setup, where all model parameters are updated during training. The fine-tuning spans 30 epochs, using text-image training datasets derived from the three data augmentation scenarios. After each epoch, performance is evaluated on the validation datasets, and loss metrics are recorded for both training and validation phases, providing insight into model learning progress as illustrated in Figure 4.

The AdamW optimizer [25] is applied with a cosine learning rate decay [26], starting at $1 \times 10^{-6}$, and no warm-up steps. Training is conducted with a batch size of one, utilizing mixed precision (FP16) to enhance computational efficiency. The model is optimized using its default cross-entropy loss function.

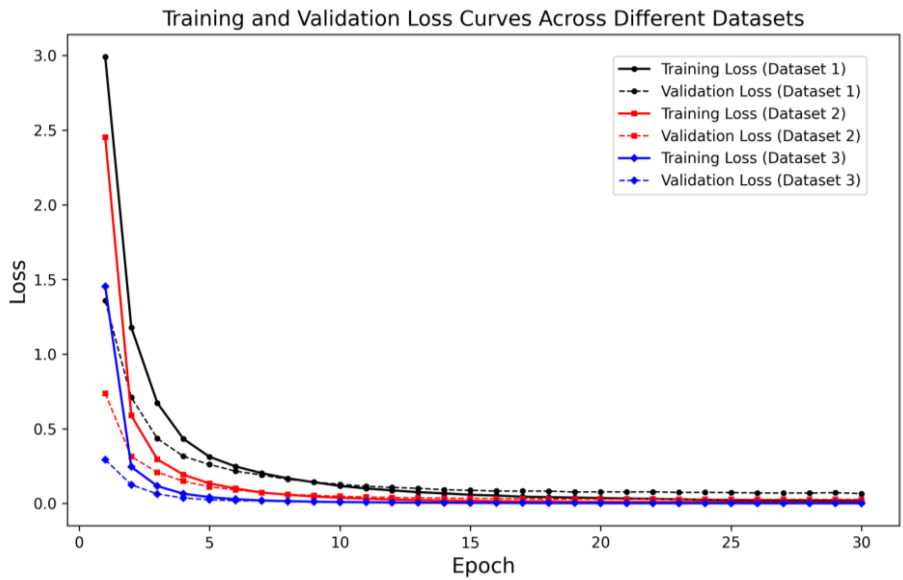

**Fig. 4.** Training and validation loss curves for the three datasets. The consistently decreasing loss across all datasets demonstrates steady model improvement over time.

## 4.3 Model Inference Process

For GPT-4o and Claude-3.5-Sonnet, inference is performed in zero-shot mode, where the models process data without any domain-specific fine-tuning. This setup leverages the models' pre-trained knowledge to directly analyze the drawing images. Prompts structured to identify GD&T symbols, tolerances, and datums guide each model in generating outputs in a structured JSON format, which is essential for consistent and straightforward comparison with the ground truth annotations. The extracted GD&T details are then parsed for evaluation.

For Florence-2, inference is conducted using the three fine-tuned models, each evaluated on its corresponding validation dataset. Prompts for each image-query pair instruct the models to output GD&T information, focusing on maintaining the correct order of GD&T elements and ensuring strict adherence to JSON formatting for comparison accuracy. To ensure consistency with the ground truth data, initial outputs undergo post-processing using *GPT-4o-mini* to correct any formatting inconsistencies. The final processed outputs are saved as JSON files, each corresponding to the index of an image for structured evaluation.

## 4.4 Evaluation Metrics

The models are evaluated using the following metrics [27], specifically tailored for the GD&T extraction task:

$$Precision = \frac{TP}{TP + FP} \tag{1}$$

$$Recall = \frac{TP}{TP + FN} \tag{2}$$

$$F1\ score = 2 \times \frac{Precision \times Recall}{Precision + Recall} \tag{3}$$

$$Hallucination = 1 - Precision = \frac{FP}{TP + FP} \tag{4}$$

where True Positives (TP) represent the number of correctly predicted GD&T key-value pairs that match the corresponding entries in the ground truth. An exact match is required for both the GD&T key (geometric characteristic, tolerance, datum reference) and its associated value to be considered a true positive. False Positives (FP) are key-value pairs predicted by the model that do not appear in the ground truth, indicating erroneous extractions or misclassifications. False Negatives (FN) refer to key-value pairs present in the ground truth that the model fails to predict. The hallucination metric measures the rate of incorrect extractions, highlighting the model's tendency to produce erroneous outputs that are not grounded in the actual data.

## 5 Results and Discussion

The three VLMs are evaluated on the validation datasets using four key evaluation metrics. GPT-4o and Claude-3.5-Sonnet serve as baseline models in a zero-shot setting, while Florence-2 is assessed across three experiments (Exp-1, Exp-2, and Exp-3) using datasets augmented with one, two, and four queries per image, respectively.

As shown in Table 1, GPT-4o achieves higher precision (59.03%) compared to Claude-3.5 (44.01%), but with lower recall (25.39%) and F1-score (35.51%). In contrast, Claude-3.5 demonstrates better recall (37.27%) but has the highest hallucination rate (55.99%) among the models. This suggests that GPT-4o is more accurate but struggles to detect as many relevant GD&T instances as Claude-3.5. The trade-off between precision and recall in zero-shot models highlights their limitations for specialized tasks like GD&T extraction, where detecting a wide range of features is critical. In such cases, high precision is important to minimize incorrect predictions, while high recall is crucial to avoid missing relevant GD&T information.

Florence-2 exhibits consistent improvements across all metrics, with the most significant gains observed in Exp-3. In Exp-1, there is an increase of 2.91% in precision compared to GPT-4o, but recall is 13.25% lower than Claude-3.5, indicating that the initial data augmentation did not significantly improve recall. In Exp-2, both precision and recall surpass the baseline models, with recall increasing by 2.71%, suggesting that additional queries help capture more GD&T features. Exp-3 delivers the best results, with precision improving by 29.95%, recall by 37.75%, and F1-score by 52.40% compared to the best baseline metrics. These results indicate that Florence-2 not only makes more accurate predictions but also captures a broader set of relevant GD&T instances, which is crucial for minimizing both false positives (FP) and false negatives (FN) in GD&T extraction.

Moreover, the hallucination rate decreases consistently across all experiments, with the largest reduction of 43.15% in Exp-3. Reducing hallucination is particularly important in GD&T extraction, as incorrect entries can lead to significant errors in quality control. These results demonstrate that Florence-2, especially in Exp-3, outperforms the baseline models across all metrics. The improvements highlight the value of data augmentation and fine-tuning, allowing a smaller, open-source VLM like Florence-2 to surpass advanced closed-source models such as GPT-4o and Claude-3.5-Sonnet. This makes Florence-2 more effective and task-specific for automating GD&T extraction in manufacturing processes, particularly for complex engineering drawings.

**Table 1.** Evaluation metrics for the three selected VLMs. Underlined values indicate the best-performing baseline closed-source model (GPT-4o or Claude-3.5) for each metric. The percentage changes for Florence-2 are shown relative to the best baseline values, with Exp-3 achieving the best performance.

| Models | Precision (%) | Recall (%) | F1 score (%) | Hallucination (%) |
|---|---|---|---|---|
| *GPT-4o* | <u>59.03</u> | 25.39 | 35.51 | <u>40.97</u> |
| *Claude-3.5-Sonnet* | 44.01 | <u>37.27</u> | <u>40.36</u> | 55.99 |
| *Florence-2 (Exp-1)* | 60.75 (+2.91%) | 32.33 (-13.25%) | 42.20 (+4.56%) | 39.25 (-4.19%) |
| *Florence-2 (Exp-2)* | 71.74 (+21.53%) | 38.28 (+2.71%) | 49.92 (+23.69%) | 28.26 (-31.02%) |
| *Florence-2 (Exp-3)* | **76.71 (+29.95%)** | **51.34 (+37.75%)** | **61.51 (+52.40%)** | **23.29 (-43.15%)** |

Figure 5 illustrates the evaluation metrics for Florence-2 (Exp-3) on the validation dataset. The results show that as the number of GD&T entries per image increases, the model's performance tends to decline, with recall and F1-score decreasing and hallucination increasing. This trend indicates that more complex drawings present greater challenges for accurate GD&T identification.

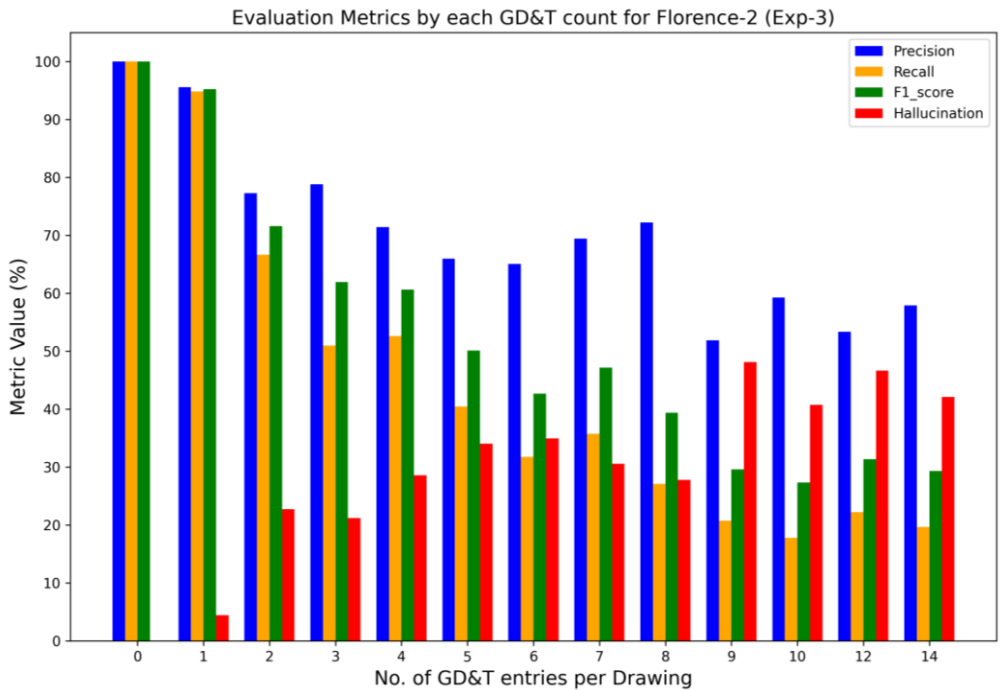

**Fig. 5.** Evaluation metrics for Florence-2 (Exp-3) on the validation dataset, indicating that as the number of GD&T entries per drawing rises, recall and F1-score generally decrease while hallucination increases, reflecting the model's sensitivity to drawing complexity.

## 6 Conclusion

This paper introduces a novel and computationally efficient method for automating GD&T information extraction from 2D engineering drawings by fine-tuning the open-source VLM, Florence-2. Despite its relatively small size of 0.23 billion parameters, Florence-2 outperforms larger closed-source models like GPT-4o and Claude-3.5-Sonnet across all evaluation metrics. The full-parameter fine-tuning approach applied to Florence-2 leads to a 29.95% increase in precision, a 37.75% improvement in recall, a 52.40% increase in F1-score, and a 43.15% reduction in hallucination compared to the best-performing closed-source baseline model. These results demonstrate that smaller, open-source models can be fine-tuned for domain-specific tasks to achieve highly accurate results, while also offering the benefit of lower computational costs associated with training.

The key contributions of this study include the creation of a structured, annotated GD&T-specific dataset, the development of data augmentation techniques tailored to this domain, and the successful fine-tuning of an open-source VLM. Tailored evaluation metrics provide a comprehensive assessment of model performance. Future work will focus on expanding the training dataset and exploring advanced data augmentation techniques and fine-tuning configurations to further enhance model performance. Additionally, fine-tuning other state-of-the-art open-source VLMs may yield even better results to extract GD&T information for downstream manufacturing tasks such as process planning, tool selection, part classification, assembly validation, and quality control.

## Acknowledgements

This work is supported by the Agency for Science, Technology and Research (A*STAR), Singapore, through the RIE2025 MTC IAF-PP grant (Grant No. M22K5a0045). It is also supported by the Singapore International Graduate Award (SINGA) (Awardee: *Muhammad Tayyab Khan*), funded by A*STAR and Nanyang Technological University, Singapore.